\pdfoutput=1

\documentclass[sigconf]{aamas} 

\usepackage{balance} 

\usepackage[T1]{fontenc}
\usepackage{graphicx}
\usepackage{amsmath}
\usepackage{amsfonts}
\usepackage{xcolor}
\usepackage{xspace}
\usepackage{subcaption}
\usepackage{pifont}

\usepackage{amsthm}
\usepackage{mathtools}

\usepackage{amsmath}

\setcopyright{ifaamas}
\acmConference[AAMAS '23]{Proc.\@ of the 22nd International Conference
on Autonomous Agents and Multiagent Systems (AAMAS 2023)}{May 29 -- June 2, 2023}
{London, United Kingdom}{A.~Ricci, W.~Yeoh, N.~Agmon, B.~An (eds.)}
\copyrightyear{2023}
\acmYear{2023}
\acmDOI{}
\acmPrice{}
\acmISBN{}

\acmSubmissionID{???}

\title[Presenting Multiagent Challenges in Team Sports Analytics]{Presenting Multiagent Challenges in Team Sports Analytics}
\subtitle{Blue Sky Ideas Track}

\author{David Radke}
\email{dradke@blackhawks.com}
\affiliation{%
  \institution{Chicago Blackhawks}
	\city{Chicago, IL}
	\country{USA \\ University of Waterloo}
}

\author{Alexi Orchard}
\email{alexi.orchard@uwaterloo.ca}
\affiliation{%
  \institution{University of Waterloo}
	\city{Waterloo, ON}
	\country{Canada}
}

\begin{abstract}

This paper draws correlations between several challenges and opportunities within the area of team sports analytics and key research areas within multiagent systems (MAS).
We specifically consider invasion games, defined as sports where players invade the opposing team's territory and can interact anywhere on a playing surface such as ice hockey, soccer, and basketball.
We argue that MAS is well-equipped to study invasion games and will benefit both MAS and sports analytics fields.
Our discussion highlights areas for MAS implementation and further development along two axes: short-term in-game strategy (coaching) and long-term team planning (management).
\end{abstract}


\newcommand{\BibTeX}{\rm B\kern-.05em{\sc i\kern-.025em b}\kern-.08em\TeX}

\begin{document}


\pagestyle{fancy}
\fancyhead{}


\maketitle 




\newcommand*{\ourmean}[1]{\overline{#1}}

\newcommand{\oursubsub}[1]{\subsubsection{#1}}

\newcommand{\optional}[1]{\textcolor{YellowGreen}{#1}}

\newcommand{\todo}[1]{\textcolor{red}{\textbf{#1}}}

\newcommand{\done}[1]{\textcolor{orange}{\st{#1}}}

\newcommand{\timtext}[1]{\textcolor{black}{#1}} 

\newcommand{\oldtext}[1]{\textcolor{cyan}{#1}}

\newcommand{\heading}[1]{\vspace{3pt}\noindent\textbf{#1 }}

\newcommand{\mysubsection}[1]{\vspace{3pt}\noindent\textbf{#1 }}

\newcommand{\newtext}[1]{\textcolor{blue}{#1}}

\newcommand{\vitem}{\vspace{-5pt}\item}

\newenvironment{guideline}{\vspace{0pt} \noindent \hrulefill \\ \emph{\bf \textcolor{blue}{GUIDELINE:}} \it }{\\ \vspace{-5pt} \hrule}

\newcommand{\squishbegin}{
 \begin{list}{$\bullet$}
  { \setlength{\itemsep}{0pt}
     \setlength{\parsep}{1pt}
     \setlength{\topsep}{1pt}
     \setlength{\partopsep}{0pt}
     \setlength{\leftmargin}{1.5em}
     \setlength{\labelwidth}{1em}
     \setlength{\labelsep}{0.5em} 
  } 
}

\newcommand{\squishtwobegin}{
 \begin{list}{$-$}
  { \setlength{\itemsep}{1pt}
     \setlength{\parsep}{1pt}
     \setlength{\topsep}{1pt}
     \setlength{\partopsep}{0pt}
     \setlength{\leftmargin}{1.5em}
     \setlength{\labelwidth}{1em}
     \setlength{\labelsep}{0.5em} 
  } 
}

\newcommand{\squishend}{
  \end{list}  
}

\newcommand{\experiment}{\vspace*{4pt}\noindent\textbf{Experiment Setup:\hspace{0.4em}}}
\newcommand{\experimentend}{}

\newcommand{\moveup}{\vspace{-8pt}}
\newcommand{\movecaptionup}{\vspace{-20pt}}
\newcommand{\movecaptionuptab}{\vspace{-17pt}}
\newcommand{\colfigwidth}{0.90\columnwidth}


\newcommand{\btable}[1]{\begin{table}[#1] \begin{center} }
\newcommand{\etable}[2]{\end{center} \vspace{-5pt} \caption{#2} \label{#1} \vspace{-15pt}\end{table}}

\newcommand{\wbtable}[1]{\begin{table*}[#1] \begin{center} }
\newcommand{\wetable}[2]{\end{center} \caption{#2} \label{#1} \end{table*}}

\newcommand{\xfigure}[5]{\begin{figure}[#1] \begin{center} \leavevmode \epsfxsize=#2 \epsfbox{#3} \end{center} \vspace{-12pt} \caption{#5} \label{#4} \end{figure}}

\newcommand{\xfigurewide}[5]{\begin{figure*}[#1] \moveup \begin{center} \leavevmode \epsfxsize=#2 \epsfbox{#3} \end{center} \movecaptionup \caption{#5} \label{#4} \end{figure*}}

\newcommand{\yfigure}[5]{\begin{figure}[#1] \begin{center} \leavevmode \epsfysize=#2 \epsfbox{#3} \end{center} \caption{#5} \label{#4} \end{figure}}

\newcommand{\xyfigure}[6]{\begin{figure}[#1] \begin{center} \leavevmode \epsfxsize=#2 \epsfysize=#3 \epsfbox{#4} \end{center} \caption{#6} \label{#5} \end{figure}}

\newcommand{\bfigure}[1]{\begin{figure}[#1]}
\newcommand{\efigure}[2]{\vspace{-8pt} \caption{#2} \label{#1} \end{figure}}

\section{Introduction}
\label{sec:intro}




Understanding how individuals learn to coordinate and cooperate is an important problem in multiagent systems (MAS)~\cite{DafoeNature2021}.
Humans demonstrate effective teamwork in many aspects of daily life to achieve their goals, one of the best examples being in team sports.
In this paper, we advocate for using MAS to extend advanced analytics for real-world sports which, in turn, will help researchers understand how teamwork evolves, the conditions under which it succeeds, and support future multiagent research.
While there has been momentum to incorporate artificial intelligence (AI) into sports~\cite{tuyls2021game}, this paper articulates the unique position of MAS to address problems of interest in both the AI and sports communities.

In 2002, the Oakland Athletics adopted a new approach to constructing and managing baseball teams, a revolution in sports analytics hence known as Moneyball~\cite{lewis2004moneyball}: using empirical statistics as a basis for roster management.
Two decades later, most baseball teams employ a staff of analysts that support roster and strategic decisions with quantitative data~\cite{elitzur2020data}.
Baseball is classified as a ``striking game'' due to it's episodic and repetitive structure, allowing for relatively easily collected data that lends itself nicely to statistics.
However, statistics alone is unable to fully capture the complexity of ``invasion games'', defined by using a goal or hoop where attacks rely on invading opponent territory and players can interact anywhere on a playing surface (e.g., soccer, ice hockey, and basketball~\cite{ellis1983similarities}).
We argue that MAS is to invasion games what statistics is to striking games.
Invasion games are well suited for the advancement of multiagent research by providing enclosed, structured environments with an abundance of data collection.

Although there is ongoing investment in using AI in soccer analytics~\cite{tuyls2021game}, sports analytics generally has not been adopted as a typical research domain within MAS.
This paper highlights several challenges from invasion team sports that are similar to MAS research topics at various levels of complexity and time horizons.
Coaches must identify players that coordinate well together, devise team-based strategies, and best responses in-match to outperform opponents.
We draw correlations between coaching challenges and the multiagent areas of team forming, coalition formation, and non-cooperative game theory.
Managing a sports team requires long-horizon planning across multiple seasons, generalizing team performance and planning player acquisitions, facing challenges similar to agent modelling, markets, and resource allocation.
We posit that MAS and sports analytics will benefit from each other's contributions at this intersection.

While this paper does not exhaustively address multiagent challenges in sports, it rather works to make correlations between and suggests avenues for advancement in sports analytics and domains within MAS.
We highlight relevant MAS topic areas using: \ding{229}.

\section{Background and Scope}
\label{sec:background}

\begin{figure}[t]
    \centering
    \begin{subfigure}[b]{0.85\linewidth} 
        \centering
        \includegraphics[width=\linewidth]{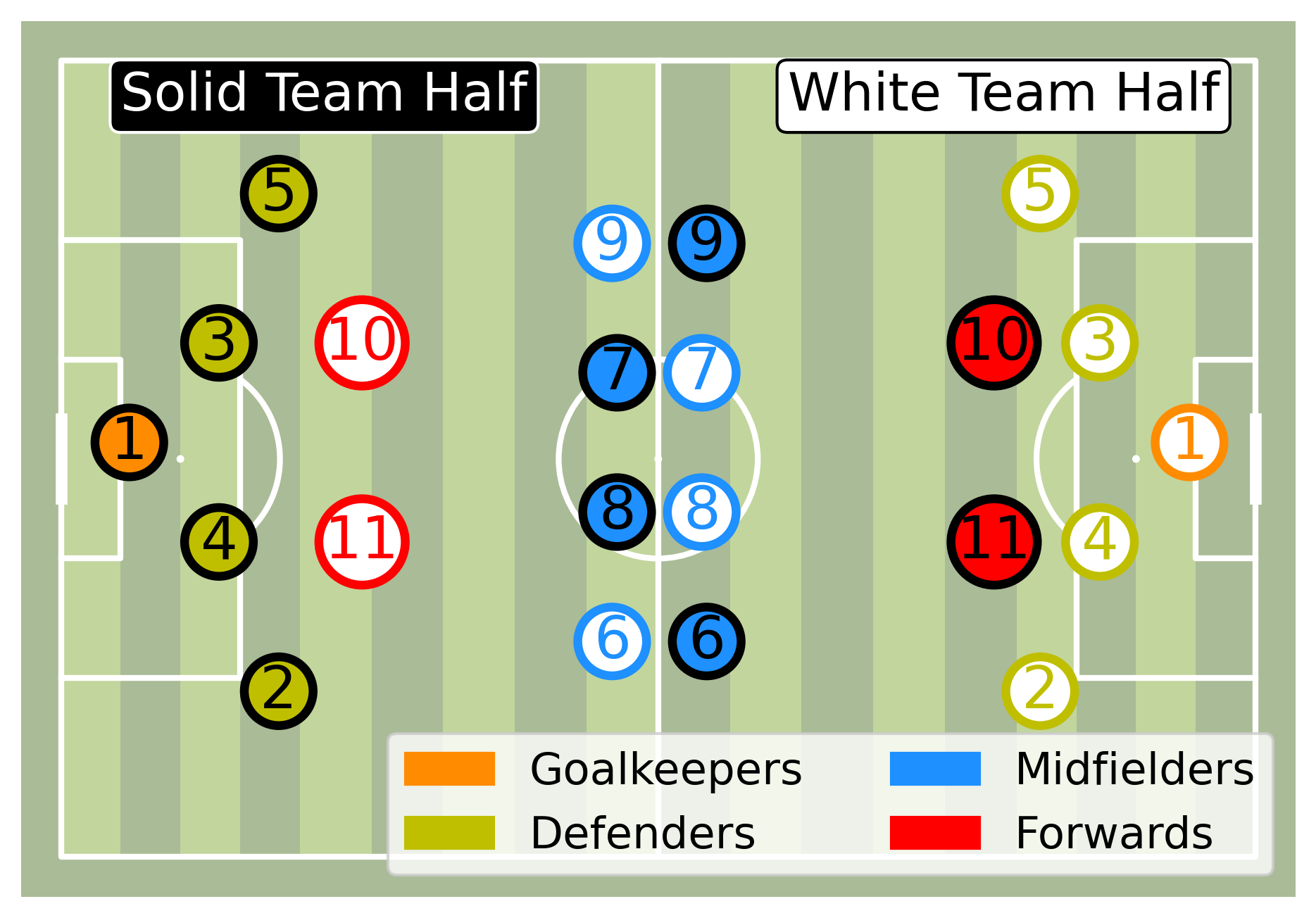}
        \caption{Soccer pitch (two teams).}
        \label{fig:soccer}
    \end{subfigure}
    \begin{subfigure}[b]{0.83\linewidth} 
        \includegraphics[width=\linewidth]{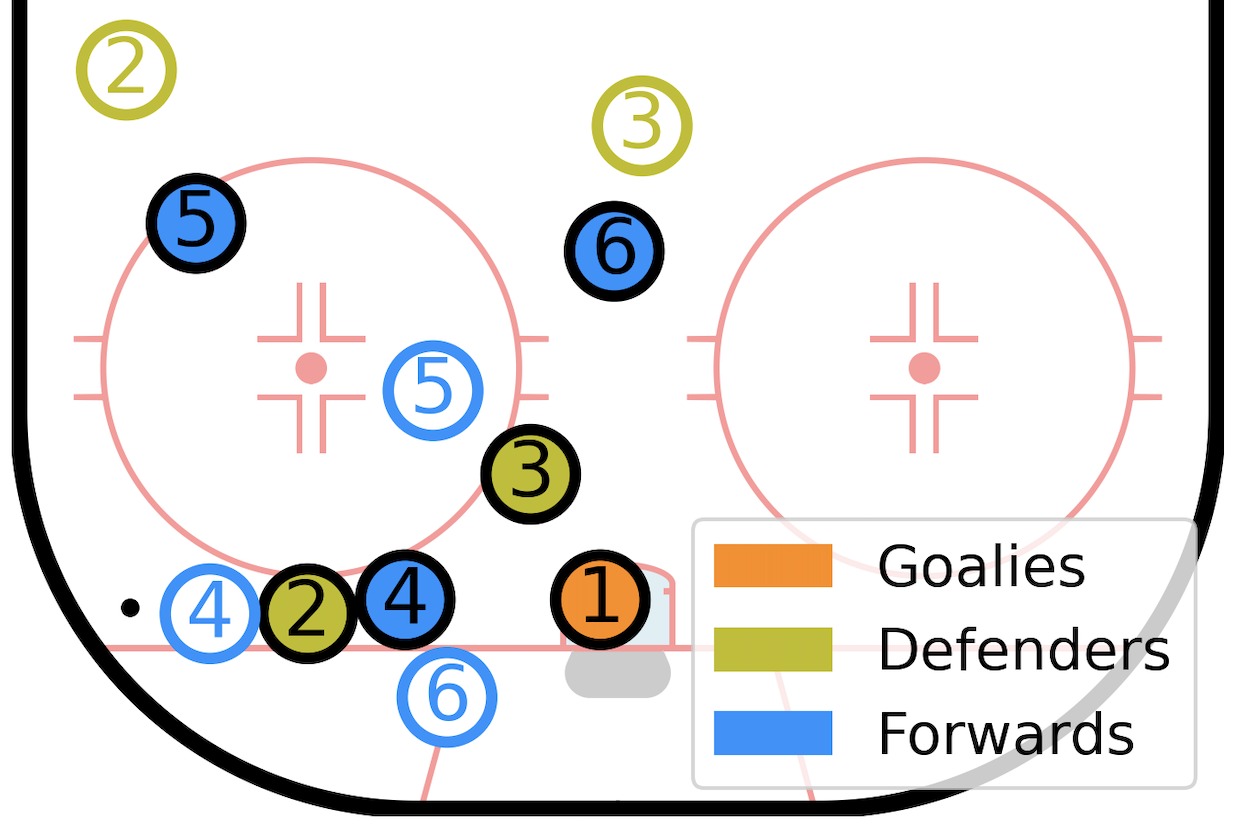}
        \caption{Ice hockey scenario (two teams) with puck.}
        \label{fig:hockey}
    \end{subfigure}
        \caption{(a) Example of a soccer (or football) pitch with two teams, and (b) an ice hockey end-zone (one-third of surface) with two teams. Players are divided into positions with individual tasks but have the overall goal of winning the game.}
        \label{fig:sports}
\end{figure}

Team sports are played in a variety of environments and governed by a diverse set of rules.
While multiagent challenges may be found in most sports, we limit the discussion of this paper to invasion games where teams of dynamically moving players participate in two-team zero-sum matches within a larger league with many teams.
This includes widely popular team sports such as soccer, ice hockey, basketball, and many more.
Matches involve two teams ($\mathcal{A}$ and $\mathcal{B}$) composed of $N$ players (agents) each.
Teams typically have substitutions, so a player might not play the entire game; thus, the set of \emph{active} players in the game at any time is $n \subseteq N$ for each of $\mathcal{A}$ and $\mathcal{B}$ (i.e., $n=6$ in ice hockey, $n=5$ in basketball, and $n=11$ in soccer).
A team can often be further divided into \emph{positions} (roles), such as defense or offense, where players within a particular position are further specialized (i.e., left and right defense).
Figure~\ref{fig:sports} shows two examples of invasion games (soccer and ice hockey) with two teams (solid and white), where players are divided into different positions.
The intermingling of players exemplifies the complexity of interactions that occur in invasion games.

We examine challenges along two axes with inter-related problem spaces: within-match coaching and team management over long horizons.
While teams aim to win individual zero-sum matches, the challenge of managing teams involves modeling a larger environment, since teams compete in leagues with many teams.
As a result, challenges at these two levels operate on different timescales, but are not mutually exclusive.
Match results impact the challenges that management faces, while management's actions impact the coaching environment.
Addressing many of the challenges in this paper requires solutions that consider impacts along both axes.



\subsection{Data Collection}


Data collected about invasion games often includes different levels of detail, from event-based records (pass, shot, or goal) to tracking player locations~\cite{rein2016big,fernandez2021framework}.
Event-based data typically records details about significant in-game events such as time, score differential, player ID, event name, and coordinates.
Thus, a collection of sequential events can be modeled as a Markov chain, where each event depends on the game state and action of the preceding state~\cite{schulte2017markov}.
Thousands of events are recorded each match; however, event-based data is unable to capture the complete context of invasion games since the positions of all players are not recorded.

Tracking data records the locations of all players on a playing surface multiple times per-second.
Tracking systems are currently deployed in the highest leagues of soccer~\cite{bialkowski2014large}, ice hockey~\cite{radke2022identifying}, and basketball~\cite{sampaio2015exploring}.
Each tracking sample records features such as timestep, player ID, spatial coordinates, and velocity for each player and the ball or puck, typically amounting to millions of data points in each game.
Tracking systems function through physical hardware on each player and ball or puck, or vision-based systems that extract detailed events or attributes such as hockey stick location and pose, a significant area of hardware and computer vision research itself~\cite{beetz2006camera,vats2022evaluating,rahimian2022optical}.
Tracking data can also be modeled as a Markov chain of events where the action space includes player movement within the playing surface and can be joined with event data to add additional event context to player movements~\cite{fernandez2021framework}.
Many sports datasets are freely available for download.\footnote{\url{https://www.kaggle.com/datasets?search=sports}}

\section{Coaching \& Match Strategy}
\label{sec:coaching}

We first discuss multiagent challenges related to sports analytics in the context of coaching.
Coaches rely on information about player, group, and opponent performances to make decisions about who plays what positions within each match.
Related multiagent problems include optimizing team arrangement, player and group valuation, and opponent prediction and modeling.
The ability to provide quantitative support to make line-up or team strategy decisions is an open problem in sports analytics.
We propose the following high-level research questions:
\textbf{RQ1}: How can multiagent research help coaches identify good combinations of players in the context of opponents and development?
\textbf{RQ2}: How can the temporal and inter-agent complexity of in-match problems within invasion games support future MAS research?



\subsection{Team Arrangement}


Understanding how to best utilize a group of players requires an understanding of multiagent interaction.
A team must operate as a cohesive group of individual players, often divided into sub-teams with various roles or positions (i.e., defense or forwards).
In sports such as ice hockey where groups of players are replaced at high frequency (i.e., players change about once per-minute), the goals of disparate groups that are used in the same position may be heterogeneous (i.e., scoring forwards versus defensive forwards).
Coaches are tasked with assessing the value of their players in continuously dynamic contexts to decide which players fit into particular roles and which sub-groups perform best together.
This is an online coalition formation problem: constructing sub-groups within their team to satisfy various types of goals conditioned on how players respond to their peers and game context~\cite{rahwan2015coalition}.

Existing coalition formation or team forming algorithms usually compose agent groups with the goal of maximizing reward, fairness, or robustness to failure~\cite{schwind2021partial}.
Agent policies in these environments are typically assumed to be static~\cite{andrejczuk2018composition}.
These algorithms could be adapted to team sports to measure some notion of expected performance; however, fully capturing the complexity of team sports for coalition formation requires considering dynamically changing environment states, opponent policies, and under-specified goals.
For example, groups that are desired to maximize defensive play (instead of offense) must consider opponent strength and properly allocate credit to defensive actions.
Future algorithms will need to understand the value of different actions and be generalizable to different rosters, opponents, and goals.

\noindent
\ding{229} Coalition Formation \\
\ding{229} Teamwork and Team Forming

\subsection{Player and Group Valuation}
\label{sec:state_value}

Traditional components of player valuation fail to consider full contributions players make to team success (i.e., mostly offensive metrics, even for defensive players).
Newer approaches have focused on learning the value of passes, carries, shots, or full trajectories, providing more context to the decision making of players or groups~\cite{beal2020learning,Radke2021Passing,fernandez2021framework,ritchie2022pass}.
However, agents' abilities are often a direct consequence of their teammates' impacts on the game state~\cite{goes2019not}.
While all player movement and locations are easily captured with tracking data, learning the value of inter-agent and off-ball/puck movement is an active area of research~\cite{spearman2018beyond,raabe2022graph}.
We emphasize that players must be evaluated not only by their individual actions, but in the context of their team, opponents, and game situations.

Learning the value of a player within a larger context requires a rich understanding of high-dimensional inter-agent interactions and causal reasoning (i.e., off-ball/puck movements).
Estimating the marginal value of agents in AI is important for credit assignment, group cohesion, and developing effective joint policies.
Some existing models use a supervised component to learn marginal contribution~\cite{rashid2018qmix}, while others rely on methods from cooperative game theory~\cite{derks1993shapley,Yan2020EvaluatingAR}; however, they mainly function in online settings.

For multiagent methods that estimate marginal contribution to be effective in sports analytics, they must perform offline policy evaluation from data.
Models must also consider the potential for different players to develop better joint policies if they were to play together.
This means understanding player types and contributions from data separate from easily identifiable signals such as scoring goals; for instance, identifying skilled defensive players.
Identifying alternative goals and multiple tasks, such as beneficial state-action pairs which may not produce reward themselves, is already an important research problem in sparse reward and offline environments~\cite{arjona2019rudder}, although mostly explored in single-agent scenarios.
Developing new AI solutions in this direction requires modeling poorly specified rewards, types of goals, and action causal/counterfactual reasoning with multiple agents.
Through this development of new AI algorithms, credit assignment and value decomposition models will be directly improved to better estimate marginal value.
Relaying this information to coaches would have a direct impact on team strategy, groupings, and group agency or joint policies.

\noindent
\ding{229} Learning Agent Capabilities \\
\ding{229} Emergent Behavior 

\subsection{Opponent Prediction and Strategy}
\label{sec:opponent_modeling}

Sports teams execute team-based strategies and systems in an attempt to outperform opposing teams.
Predicting opponent team formations and strategies is a challenging problem typically delegated to coaching staffs.
Providing quantitative support for this challenge requires a rich understanding of agent-based modeling, prediction, coordination, and identifying tendencies~\cite{visser2000recognizing,lucey2012characterizing}.
These problems are reminiscent of opponent modeling and planning, such as work done within the popular multiagent RoboSoccer domain~\cite{pourmehr2011overview,stone2000defining,ledezma2009ombo}.
Further AI development in this direction must consider multiple layers of complexity such as game state, individual player match-ups, player availability, and team-wide risk.

There may also be situations where an opponent's individual incentives may diverge from their team-based strategy, causing their behavior to stray from their team's strategy based on team alignment and types of goals~\cite{radke2022importance}.
Learning when this may happen requires a rich understanding of agent and team-level modeling around game theory~\cite{Radke2022Exploring}, bounded rationality~\cite{simon1990bounded}, and group alignment~\cite{schroder2016modeling} beyond the current state of those fields.
Utilities and reward-based incentives may not be enough to describe the behavior of players depending on game or group context.
Progress on individual and team-level opponent modeling and strategy prediction will push multiagent research to develop better models of human behavior and strategic decisions within domains with high stochasticity.
These steps forward will directly benefit other multiagent domains such as autonomous vehicles~\cite{fisac2019hierarchical} and strategic reasoning under uncertainty~\cite{van2005logic}.


\noindent
\ding{229} Agent-based Modelling and Simulation: Applications/Analysis
\ding{229} Non-Cooperative Game Theory

\section{Management \& Long-Term Strategy}
\label{sec:management}


Management considers both short-term performance with higher-level constraints and long-term planning, resulting in multi-tiered problems that are particularly interesting from an AI perspective.
Roster analysis, roster construction, and financial strategy are areas with different, but related, inter-dependent multiagent challenges.
To the best of our knowledge, using sports to develop and evaluate AI models related to financial strategies has not been explored.
We propose the following high-level research questions:
\textbf{RQ1}: How can multiagent research support management to compose a cohesive team and construct environments that promotes player development under financial restrictions?
\textbf{RQ2}: How can team management inspire multiagent research surrounding group valuation while considering economic strategies and diverse opponents?

\subsection{Roster Analysis}
\label{sec:roster_analysis}

Management conducts recruiting, drafting, trading, and signing players with the overall objective of creating a cohesive and high performing team.
When acquiring players, management needs to accurately analyze the current state of their roster to identify areas which could be improved or those that may be overvalued.
Analyses include the structure, cohesiveness, and playing style of their current team, as well as predicting the future developmental trajectories of their prospects~\cite{schuckers2011s}.
Playing style refers to the personality or characteristics of a player, group, or entire team (i.e., defensive forwards vs. offensive forwards).
Playing style is easy for experts to define; however, is not currently easily extractable from data.

Some prior work has focused on identifying groups of players with high chemistry or performance~\cite{liemhetcharat2015applying,Ljung2018PlayerPV} or learning semantic representations of players from event data~\cite{liu2020learning}.
Analyzing playing style may be similar to the thread of research studying agent types in ad hoc teamwork~\cite{Albrecht2017ReasoningAH} or role diversity in multiagent reinforcement learning (RL)~\cite{le2017coordinated,hu2022policy,Radke2022Exploring}.
Understanding how players' styles evolve and are impacted by their environment (teammates and usage) is an area for future research that requires AI models that identify characteristics of group agency, sub-group joint policy development, and emergent role specialization.
This direction will require new advancements in multiagent offline RL, multiagent inverse RL, behavior cloning, and offline policy evaluation to better understand the developmental impacts of team structure and multiagent interaction on agents' policies.
These advancements will push multiagent research to further understand group agency, joint-policy evaluation, and how policies are influenced by surroundings.


\noindent
\ding{229} Learning Agent Capabilities \\
\ding{229} Modelling and Simulation of Societies

\subsection{Roster Construction}
\label{sec:roster_construction}



A key challenge of management is constructing a team that can generalize to various opponents throughout a season.
Successful teams are typically composed of players with heterogeneous and complimentary skills, placing emphasis not only on analyzing a roster, but planning for how to construct a team through drafting, developing, trading, or signing free agents.
A draft is when teams in a league select from a pool of prospective players to claim their rights, comparable to selecting objects out of a pool of items by agents with different preferences in game theory~\cite{bouveret2014manipulating}.
Draft strategies have been studied from a game theoretical perspective~\cite{brams1979prisoners} analyzing different types of utility functions based on a team's needs.
Whereas roster analysis can help identify areas for improvement (Section~\ref{sec:roster_analysis}), predicting opponents' draft strategies and a best response is an interesting area of future work that requires a rich understanding of game theory and multi-level planning.

Maximizing team performance is not simply about maximizing projected utility since group performance depends on players' abilities to develop and work together.
Successful teams solve problems of anticipation, distributed intelligence, and theory of mind to work as a collective organism~\cite{williamson2014distributed,bransen2020player}.
Teams can evolve to perform greater than the sum of individual parts~\cite{williamson2014distributed}, much like how groups in multiagent RL or evolutionary game theory develop complimentary policies by training together~\cite{Radke2022Exploring}.
However, the scenarios that allow agents to develop methods to best work together are still not fully understood in multiagent research~\cite{Durugkar2020BalancingIP,radke2022importance}.

Similar to problems in roster analysis, management's drafting or player acquisition strategies may shift depending on the current or projected composition of their team.
While understanding types of agents that form chemistry has been an active area of sports analytics research~\cite{liemhetcharat2015applying}, further development in the multiagent context is required to consider long-term team strategies, planning, and group alignment or incentives.
These problems will push the MAS community to better understand how agents form joint policies.

Lastly, losing players from a roster due to player injuries, retirement, or free agency is a common occurrence in team sports which often challenge a team's robustness at certain positions.
Team forming with an emphasis on being robust to failures or outages is a common problem in MAS~\cite{schwind2021partial} that can directly support how managers construct their team.
Providing insight into different degrees of group performance and robustness will provide management with more information when accumulating risk.
Future MAS work on group robustness to failures must expand to consider long-term temporal projections of agent value, uncertainty about agent value, and contractual agreements of various lengths.

\noindent
\ding{229} Markets, Auctions, and Non-Cooperative Game Theory \\
\ding{229} Teamwork and Team Formation 

\subsection{Economic Strategies}


Managing a professional sports team is inherently coupled with economic strategies.
Players are given salaries and monetary incentives based on their performance; however, management is often constrained by a salary cap or luxury tax, limiting the amount of capital a team can allocate.
Further constraints include contract term limits or the percent of salary cap allocated to one player.

Operating within these constraints forces management to perform resource allocation while planning for future versions of their roster.
Rational behavior suggests players will accept the most competitive salary they are offered, while teams wish to offer as little capital in consideration of other negotiations.
The interaction between a player and team is similar to a 2-player donation game~\cite{Santos2021SocialNI}.
However, players can receive offers from other teams, increasing the complexity of the strategy space to create a particularly interesting domain for behavioral game theory.

A challenge for management beyond agreeing to a single player's contract is how to allocate a team's resources to actually acquire the types of players they identify through roster analysis and construction.
Salaries across sports have been shown to be consistent with a player's performance~\cite{garner2016business}; thus, a team trying to acquire all of the best performing players may run out of available resources or experience diminishing returns on investments.
Fairly allocating capital to players based on their marginal contribution can be modeled as resource allocation problem across $N$ agents, although the value of $N$ may not be well defined.
Contracts typically last for a variable number of years, meaning strategies for allocating resources needs to plan for longer time horizons.
The financial and strategic challenges of constructing and maintaining a sports team create several dimensions of challenging multiagent problems.
This provides an interesting domain to support further multiagent research where game theoretical incentives are dynamic with opponent team strategies, roster composition, and contract landscape.
Strategic analyses that inform any aspect of financial decisions must properly model downstream impacts on other areas of management and the behavioral incentives of other teams or players.

\noindent
\ding{229} Fair Allocation \\
\ding{229} Markets, Auctions, and Non-Cooperative Game Theory




\section{Conclusion}
\label{sec:conclusion}

This paper identified several challenges within invasion game sports analytics that are related to main research problems across MAS. 
We argue that MAS is to invasion games what statistics is to striking games.
Most team decisions are currently made using domain knowledge; however, the ability to provide quantitative support to these decisions is a promising real-world domain for future multiagent research.
Addressing many of these challenges will require further advancement in multiagent AI to better capture the full complexities of value decomposition, opponent modeling, and resource allocation along multiple time horizons.


\begin{acks}
This research is funded by the Natural Sciences and Engineering Research Council of Canada (NSERC), an Ontario Graduate Scholarship, and the University of Waterloo President's Graduate Scholarship.
We also thank Tim Brecht from the University of Waterloo and Jeff Greenberg, Sam Forstner, Mark Weinstein, Albert Lyu, Chris McCorkle, and Ryan Kruse from the Chicago Blackhawks for their feedback and useful discussion on earlier drafts of this work.
\end{acks}

    
    



\bibliographystyle{ACM-Reference-Format} 


\end{document}